\documentclass[conference]{IEEEtran}
\IEEEoverridecommandlockouts
\usepackage{cite}
\usepackage{amsmath,amssymb,amsfonts}
\usepackage{algorithmic}
\usepackage{graphicx}
\usepackage{textcomp}
\usepackage{xcolor}
\usepackage{url}

\usepackage{authblk}

\def\BibTeX{{\rm B\kern-.05em{\sc i\kern-.025em b}\kern-.08em
    T\kern-.1667em\lower.7ex\hbox{E}\kern-.125emX}}
\begin{document}

\title{ToW3D: Consistency-aware Interactive Point-based Mesh Editing on GANs
\thanks{* Corresponding author.}
}

\author[1]{Haixu Song}
\author[1]{Fangfu Liu}
\author[2]{Chenyu Zhang}
\author[1,*]{Yueqi Duan}

\affil[1]{Department of Electronic Engineering, Tsinghua University, Beijing, China}
\affil[2]{Department of Automation, Tsinghua University, Beijing, China}
\affil[ ]{\texttt{\{shx22, liuff23, cyzhang21\}@mails.tsinghua.edu.cn;duanyueqi@tsinghua.edu.cn}}
\maketitle

\begin{abstract}
In this paper, we propose ToW3D that enables precise and consistent control over 3D generative adversarial networks (GANs) with the Tug-of-War competition between shape deformation and appearance consistency. Existing point-based GAN editing methods such as DragGAN and GANWarping have yielded impressive performance for 2D image manipulation. However, as 3D generators present weaker generalization ability compared with 2D due to limited training data, they would suffer from drastic changes in global appearance when editing local areas of meshes. To address this, we design a pipeline of ``drag locally, shove globally'', which iteratively performs two optimization steps: 1) pull the point towards the target, and 2) push the structure and semantics back to the source. Specifically, we design a structure adaption module based on structure which guarantees the preservation of basic geometric properties, and a semantic preservation module that maintains semantic similarity across different views. Extensive qualitative and quantitative experiments demonstrate superiority of our ToW3D approach over prior methods in terms of appearance consistency and fidelity especially under large deformations.
\end{abstract}
\section{Introduction}
\label{sec:intro}

High-quality 3D models play a crucial role in diverse industries, including applications in gaming, robotics, and virtual reality. Nevertheless, the manual creation of extensive 3D models demands a considerable investment of both time and labor. With the emergence of generative models~\cite{karras2019style}, abundant 3D generative models have come to the forefront, demonstrating their capacity to craft elaborate geometric structures, diverse topologies, and intricate texture details~\cite{gao2022get3d,liu2023sherpa3d}.

However, the process of generating a mesh through 3D generative models frequently falls short in providing precise constraints that meet the specific requirements of users. As a critical and foundational step in the geometric modeling pipeline, mesh editing has received attention in recent years~\cite{pietroni2022hex}. Conventional mesh editing methods employ regularizers to modify certain regions while preserving the fidelity of the rest shape~\cite{sorkine2004laplacian}. While these methods perform admirably under the assumption of homogeneous deformation behavior across the object, recent studies discover that they might be ineffective in practical applications involving  heterogeneous behavior~\cite{sumner2005mesh}.
Data-driven methods have been proposed to address the aforementioned problems, but they usually rely on 3D data or fail to provide precise control~\cite{park2019deepsdf}.

\begin{figure}
    \centering
    \includegraphics[width=0.5 \textwidth]{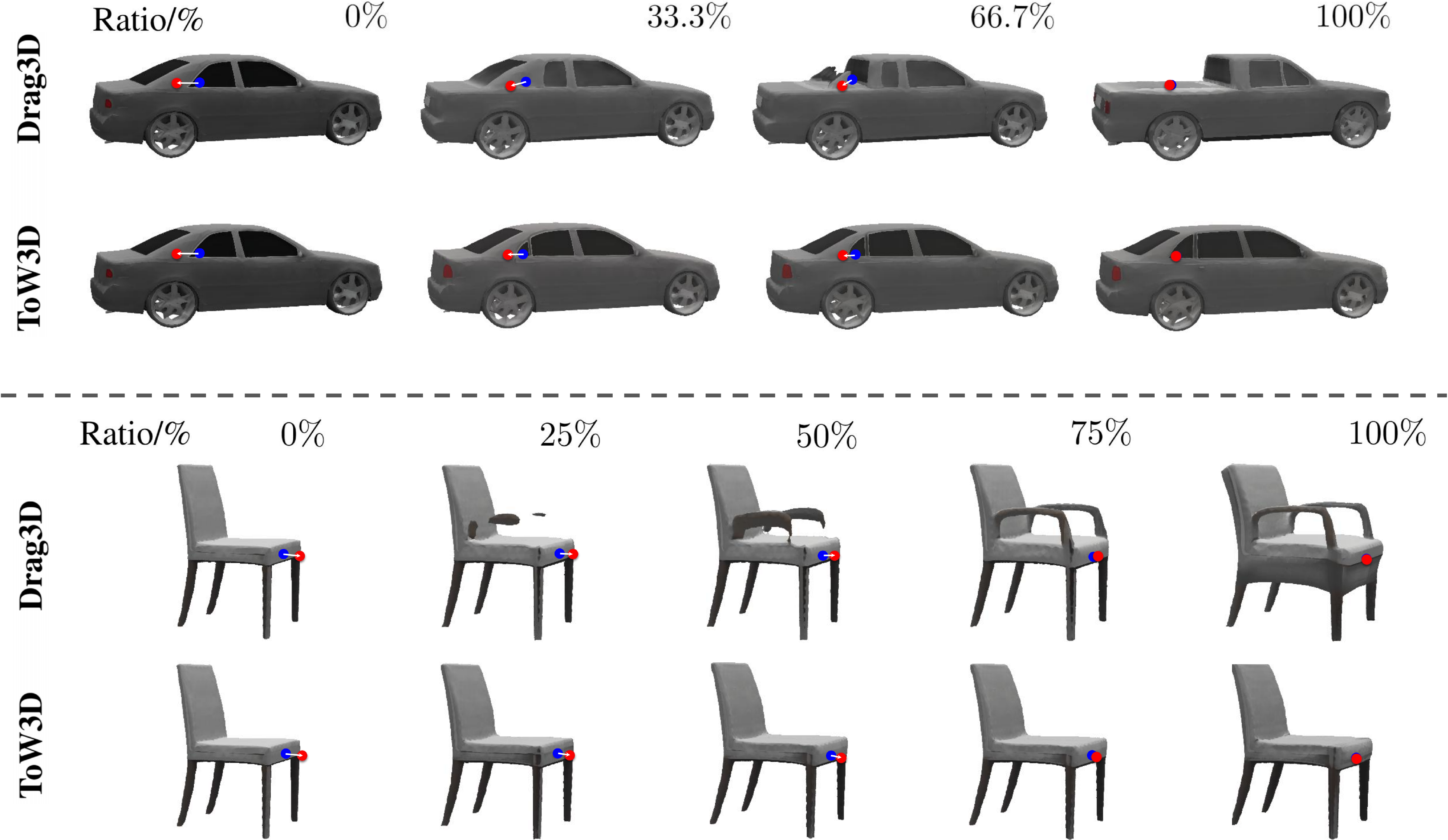}
    \caption{Qualitative comparison between ToW3D and Drag3D (an implementation of DragGAN for mesh editing). ``Ratio'' indicates the current stage of the optimization process. We observe that Drag3D induces a gradual transformation in appearance while guiding handle points (\textcolor{blue}{blue} dots) toward target points (\textcolor{red}{red} dots). In contrast, ToW3D ensures consistent appearance throughout the entire optimization process. 
    }
    \label{fig:teaser}
\end{figure}
Recently, DragGAN~\cite{pan2023drag} and GANWarping~\cite{wang2022rewriting} allow users to manually select a variable number of handle points and guide their movement towards corresponding target points. While these interactive point-based GAN editing methods have exhibited remarkable performance in the domain of 2D image editing, 3D point-based GAN editing still remains less explored. An intuitive idea is to directly extend 2D manipulation methods to 3D. However, such methods may encounter significant global appearance changes when editing local areas of meshes since 3D GANs exhibit weaker generalization ability compared with 2D due to limited 3D data~\cite{deitke2023objaverse}. As illustrated in Fig.~\ref{fig:teaser}, editing local areas may result in transforming the vehicle category into a different type or modifying the shape of a chair.

In this paper, we first propose ToW3D, a mesh editing method to manipulate a specific region of a mesh precisely while maintaining global appearance consistency. 
Motivated by the observation that the global appearance undergoes a gradual transformation throughout the process of local deformation as shown in Fig.~\ref{fig:teaser}, we design a framework with the Tug-of-War competition between shape deformation and appearance consistency to ensure the appearance consistency throughout each step. 
The framework follows the strategy of \textit{``drag locally, shove globally''} and comprises two iterative optimization steps. 
The first step focuses on directing the handle points towards the target, which may lead to slight changes in the global appearance. Subsequently, the second optimization step guides the structure and semantics to move back to the source, aligning them with the original mesh for consistency.
Specifically, we present a structure adaption module and a semantic preservation module to ensure appearance consistency.  
The structure adaption module ensures the maintenance of fundamental geometric attributes through the evaluation of the extracted structure from the mesh within the feature space. Besides, we also present a structure tracking method to address drifting problems which existing point tracking methods struggle with in texture-less areas.
Since it is challenging to directly extract semantics from the mesh, the semantic preservation module renders the mesh from multiple views and extracts high-level semantic features. Comprehensive qualitative and quantitative evaluations demonstrate that our ToW3D outperforms existing methods regarding precision and consistency especially under large deformations.

\section{Related Work}
\label{sec:relatedwork}

\textbf{3D Generation}
The rapid advancements in 2D generative models in image generation have also prompted significant progress in 3D generation~\cite{karras2019style,karras2020analyzing,karras2021alias,huang2022multimodal,ye2024dreamreward,liu2024make}. Previous methods attempt to extend 2D CNN generators directly to 3D voxel grids~\cite{wu2016learning,gadelha20173d,henzler2019escaping}, point clouds~\cite{yang2019pointflow,achlioptas2018learning,mo2019structurenet}, implicit functions~\cite{mescheder2019occupancy,chen2019learning}, or octrees~\cite{ibing2023octree} which not only lack the ability to generate textured geometry but are also incompatible with standard graphics engines. In contrast, Get3d~\cite{gao2022get3d} generate meshes with high geometric and textural details, as well as arbitrary topological structures by employing a differentiable explicit surface extraction method and a differentiable rendering technique.
Consequently, some research leverage text to control the generation of 3D objects~\cite{jain2022zero,poole2022dreamfusion}. Furthermore, contemporary research has extended the focus to include image-based control over 3D generation, resembling single-viewpoint 3D reconstruction~\cite{wang2023score,trevithick2021grf,duggal2022topologically}. 
However, regardless of whether control is employed through text or images, the generated 3D objects may not fully satisfy specific user requirements. To address this, we introduce a mesh editing algorithm to further align with individual needs and preferences.

\textbf{Image and Mesh Editing}
Researchers can manipulate unconditional GANs by modifying latent vectors. Recently, some approaches have employed fine-grained editing based on points.
GAN warping~\cite{wang2022rewriting} involves updating a GAN using warped images, but the generated images may exhibit similar warping, and their realism is not guaranteed. UserControllbleLT~\cite{endo2022user} allows for image editing but it can only utilize a single point, and there is no assurance that the handle point will reach the target point.  To address these challenges, DragGAN~\cite{pan2023drag} can simultaneously handle multiple points and guide them to their respective target points, resulting in the generation of realistic images.
Mesh editing techniques enable users to control geometric transformations by providing handles. Traditional approaches include energy-minimizing formulations~\cite{sorkine2004laplacian,sorkine2007rigid} and mesh skinning techniques~\cite{jacobson2014skinning,fulton2019latent} which may lack fine-grained control over details. 
In recent years, data-driven approaches have emerged to predict deformations. Implicit learning methods which leverage implicit occupancy grids~\cite{yumer2016learning}, neural signed distance functions~\cite{park2019deepsdf,deng2021deformed}, and radiance fields~\cite{mildenhall2021nerf,tian2023mononerf,tian2024semantic} within 3D data have gained popularity due to their robust generative capabilities.
There also exists some works addressing the scarcity limitation of 3D data with extensive 2D data~\cite{hyung2023local,mikaeili2023sked}. 
However, these methods do not offer direct geometric control and cannot guarantee precision in the deformation process.

\section{Method}
\label{sec:method}
\begin{figure*}
    \centering
    \includegraphics[width=1\textwidth]{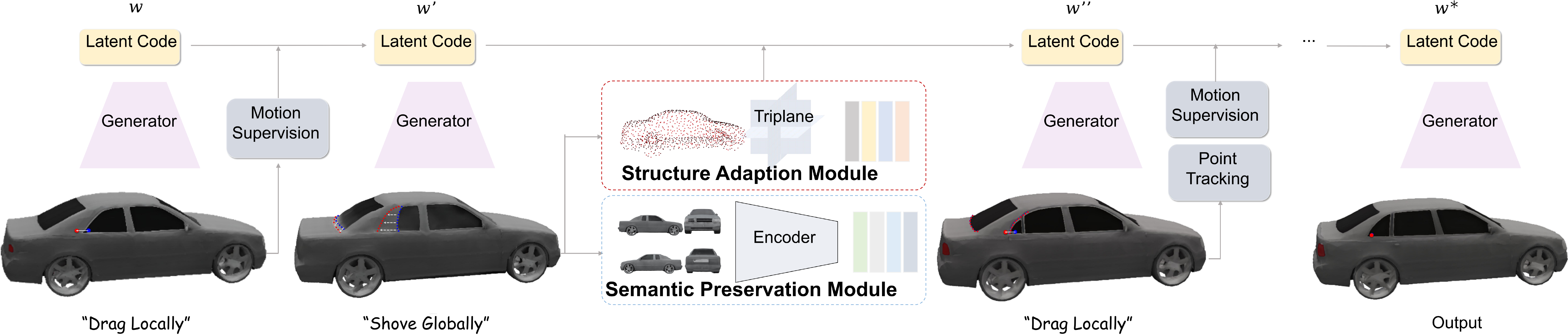}
    \caption{ 
    Overview of our proposed framework. Given a mesh generated by GANs, the user specify several handle points (\textcolor{blue}{blue} dots) and their corresponding target points (\textcolor{red}{red} dots). Our method preforms two optimization steps iteratively: the first optimization step aims to track and direct the handle points towards the target points, and it may result in appearance changes(\textcolor{blue}{blue} lines); the second optimization step encompasses structure adaption module (Section~\ref{sec:structure_adaption}) and semantic preservation module (Section~\ref{sec:semantic_preservation}) which respectively guide the structure and semantics back to the source (\textcolor{red}{red} lines) to ensure the consistency with the original mesh. The iterative process concludes upon the handle points reaching their corresponding target points.
    }
    \label{fig:pipeline}
\end{figure*}

% Our mesh editing method is designed based on , which is a popular interactive image manipulation method. 
In this section, we first review DragGAN~\cite{pan2023drag} which is a popular interactive image manipulation method and then describe our proposed ToW3D approach. Then, we introduce structure adaption module and semantic preservation module respectively.
\subsection{Preliminaries of DragGAN}
\label{sec:review of draggan}
DragGAN~\cite{pan2023drag} has demonstrated impressive performance in the field of image  manipulation recently, which consists of two key components: 
1) a feature-based motion supervision that guides handle points to move towards their target positions.  
\begin{equation}
\mathcal{L}=\sum_{i=0}^n\sum_{\boldsymbol{q}_i\in\Omega_1(\boldsymbol{p}_i,r_1)}\|\mathrm{F}(\boldsymbol{q}_i)-\mathrm{F}(\boldsymbol{q}_i+\boldsymbol{d}_i)\|_1
\end{equation}
2) a  point tracking approach that leverages the discriminative generator features to consistently localize the positions of handle points.
\begin{equation}
        \boldsymbol{p}_i:={\arg\min}_{\boldsymbol{q}_i\in\Omega_2(\boldsymbol{p}_i,r_2)}\|\mathbf{F}'(\boldsymbol{q}_i)-f_i\|_1
\end{equation}

\subsection{ToW3D}

An overview of our mesh editing pipeline is depicted in Figure~\ref{fig:pipeline}. Given a mesh $m$ generated by a GAN with latent code $w$, users can specify a set of handle points $\{p_i = (x_{p,i},y_{p,i},z_{p,i})|i=1,2,...,n\}$ and their corresponding target points $\{t_i = (x_{t,i},y_{t,i},z_{t,i})|i=1,2,...,n\}$, where $t_i$ denotes the target point for $p_i$. The primary objective is to precisely guide the specific regions represented by the handle points to reach the positions of their corresponding target points while ensuring consistency in the overall global appearance.

Given the user's input, we define the mesh editing process as an optimization problem. As shown in Figure~\ref{fig:pipeline}, we perform two optimization steps iteratively by following the strategy \textit{``drag locally, shove globally''} which indicates the Tug-of-War competition between shape deformation and appearance consistency.

The first optimization step comprises two sub-steps: point tracking to update handle points in the edited mesh and motion supervision which employs a loss to guide the handle points to move towards their target points similar to DragGAN~\cite{pan2023drag}. Given a latent code $w$, we obtain a new latent code $w'$ after motion supervision sub-step in the first optimization step, along with a new mesh  $m'$ generated by $w'$. It is worth noting that this update may introduce slight appearance changes not confined solely to the targeted area for editing compared to the original mesh $m$ generated by $w$.
Therefore, the second optimization step aims to align appearance consistency with the original mesh by guiding the geometric and semantics back to the source respectively. 

We propose the structure adaption module to ensure the preservation of fundamental geometric attributes through the optimization process. We sample structure points in the initial mesh vertices set which represent basic geometric attributes of the mesh.
We then devise a structure tracking method to localize structure points within the updated mesh $m'$ instead of directly tracking each structure point since conventional point tracking methods encounter drifting problems in texture-less areas. We further employ a loss to assess the similarity of the structures, aiming to enforce the adaption of geometric.

Furthermore, it is crucial to maintain the semantics of the mesh during the editing process and we introduce a semantic preservation module to achieve this. Since it is challenging to extract semantics from meshes directly, we utilize a differentiable render technique to generate images across multiple views. We define a loss that quantifies the similarity between the original and edited mesh in the semantic space. After the second optimization step, we get a new latent code $w''$ and corresponding new mesh.

Finally, we track the handle points in the updated mesh as illustrated in the first optimization step. Once the handle points are tracked, we repeat the aforementioned optimization steps based on the new handle points and latent code. This iterative optimization process continues until all the handle points reach the positions of their corresponding target points. Typically, this process requires 30-200 iterations in our experiments. However, users have the flexibility to stop the optimization at any intermediate step. 
After the editing process, users can input new handle and target points to further edit the mesh until they are satisfied with the results. 
\begin{figure*}
    \centering
    \includegraphics[width=0.9 \textwidth]{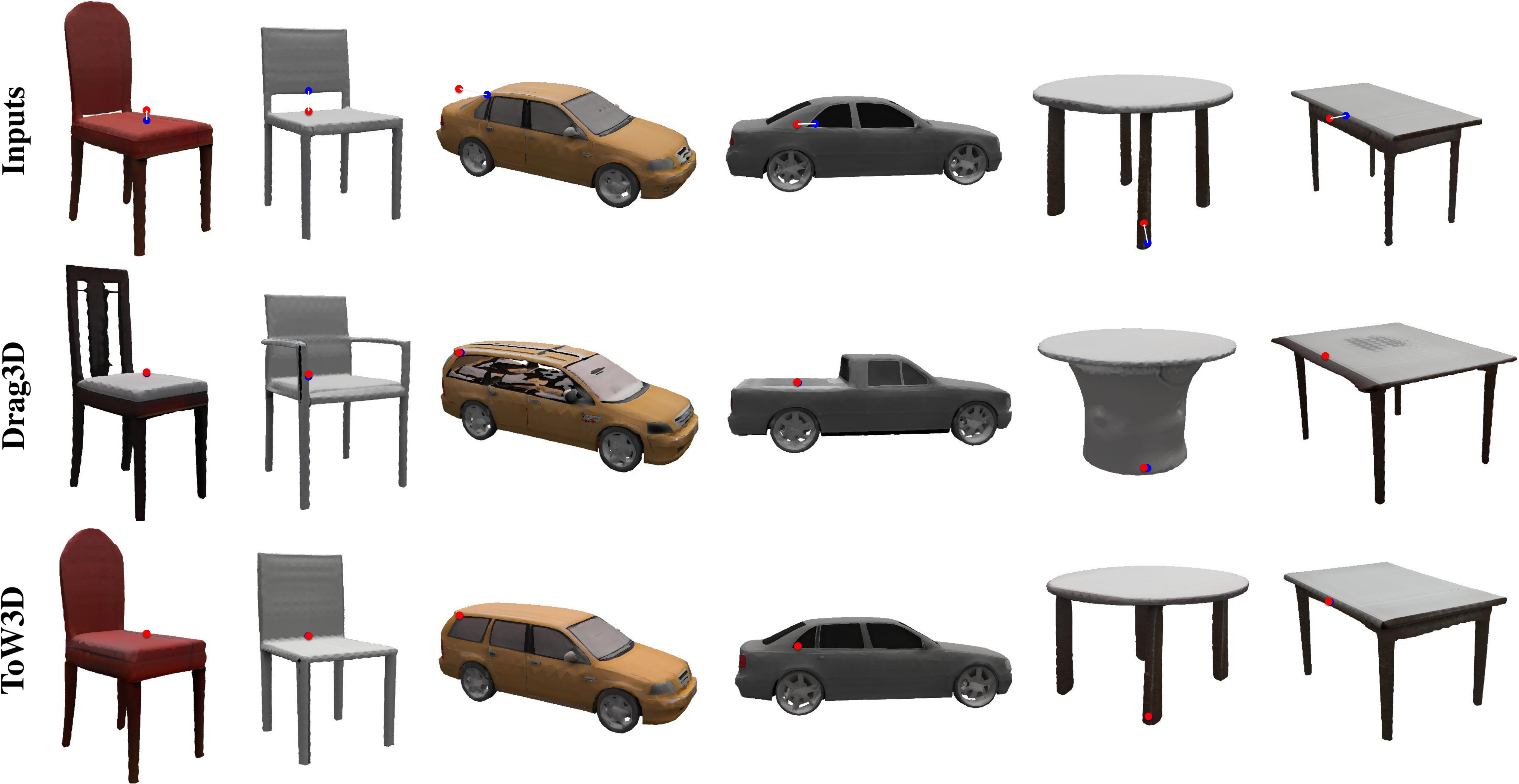}
    \caption{Qualitative comparison of our ToW3D to Drag3D on the task of directing handle points (\textcolor{blue}{blue} dots) to target points (\textcolor{red}{red} dots). Our approach demonstrates superior precise and consistent results across diverse categories.
    }
    \label{fig:qualitative_comparison}
\end{figure*}
\subsection{Structure Adaption Module}
\label{sec:structure_adaption}

To ensure basic geometric properties, we propose structure adaption module. We first uniformly sample a set of structure points in the initial mesh vertices set as a representation of fundamental geometric attributes of mesh $\{s_i = (x_{s,i},y_{s,i},z_{s,i})|i=1,2,...,m\}$. 
Each motion supervision sub-step induces slight movement of structure points and we need to track these structure points. 
Although the feature spaces of GANs are so discriminative as to facilitate effective motion tracking through nearest neighborhood search in the feature space, such point tracking method suffers from drifting issues within texture-less areas due to feature similarity. Since structure points may appear in such texture-less area, developing a new tracking method is necessary to mitigate drifting issues. Consequently, instead of tracking individual points, we track the entire structure, leveraging the predictive capacity of points within texture-less areas based on the overall structure. Since it is not feasible to assume the variation of structure as simple rigid transformation, we reformulate the structure tracking problem as an optimization problem. We define an energy function to measure the structure variation as in (\ref{equ:energy}):
\begin{equation}
\label{equ:energy}
    \mathcal{L}_{st}=\sum_i\sum_{j\in N(i)}\|(s_i^{'}-s_j^{'})-R_i(s_i-s_j)\|^2
\end{equation}
where $s_i$ is the initial position of structure point, $s_i'$ is the corresponding updated position, $N(i)$ is the set of points adjacent to $s_i$, and $R_i$ is the rotation matrix for the $s_i$. 

By leveraging the structure tracking method we design, we localize the structure points in the new mesh. We supervise intermediate features of the generator based on the idea that feature of the generator should exhibit minimum variation on the same structure points. Specifically, we leverage the feature planes, we can calculate the feature vector $f_{i}^{t}\in{R}^{32}$ as in~(\ref{equ:feature}):
\begin{equation}
\label{equ:feature}
    f_{i}=\sum_{e}\rho(\pi_{e}(p_i))
\end{equation}
where $\pi_{e}(p_i)$ denotes the projection of point $p_i$ to the feature plane $e$ and $\rho(\cdot)$ is bilinear interpolation of the features.

Instead of supervising the features of each point, we supervise the features of entire structure. 

We employ a loss to assess the similarity of the structures, aiming to minimize geometric changes. The loss is defined as in (\ref{equ:graphloss}):
\begin{equation}
\label{equ:graphloss}
\mathcal{L}_{geo}=\sum_{i}\|\mathrm{\Delta F}(s_i)-\mathrm{\Delta F}(s_i')\|_1
\end{equation}
where $\mathrm{\Delta F}({s}_i)$ represents the patched feature of structure point $s_i$ instead of the point feature to reduce uncertainty. 

\subsection{Semantic Preservation Module}
\label{sec:semantic_preservation}
As the motion supervision guide handle point towards their target point, the latent code and corresponding mesh are updated. However, unexpected semantic changes may occur in the new mesh. Therefore, we propose semantic preservation module to ensure semantics consistency throughout the optimization process. 
Different from structure adaption module maintaining low-level geometric information, semantic preservation module supervises the high-level feature to ensure slight geometric change would not cause the drastic change to the semantic. Due to the fact that directly extracting semantics from 3D data is so challenging that we leverage extensive 2D prior images. We render the mesh from multiple views via differentiable rendering technique such as Nvdiffrast~\cite{laine2020modular}. We apply the pre-trained model CLIP~\cite{radford2021learning} to extract the semantics of image for each view. 
\begin{equation}
    e_{i}=\text{CLIP}(\Phi^{*}(m))\in{R}^{512}
\end{equation}
where $m$ denotes the initial mesh, $\Phi^{*}(\cdot)$ is the differential render technique and $e_i$ denotes the embedding of image render from the $i_{th}$ view.

After motion supervision which generate a new mesh, we utilize the same render technique to generate images for each view and extract corresponding semantics.
\begin{equation}
    e_{i}'=\text{CLIP}(\Phi^{*}(m'))\in{R}^{512}
\end{equation}
where $m'$ denotes the updated mesh and  $e_i'$ is the corresponding embedding of image render from the $i_{th}$ view.

Finally, we define a semantic loss to minimize the semantic consistency across multiple views as denoted in:
\begin{equation}
    \mathcal{L}_{sem}=\sum_i\cos(e_{i},e_{i}')
\end{equation}
By minimizing the semantic loss, we preserve the semantics throughout the optimization process.

\section{Experiment}

In this section, we conduct extensive experiments to demonstrate the effectiveness of our method. We first introduce the experimental settings. Then we present qualitative and quantitative comparisons with other methods. Finally, we perform ablation studies to verify the effectiveness of our proposed modules.
\subsection{Experiment Setup}
We evaluate our method based on GET3D~\cite{gao2022get3d} pretrained on the large-scale 3D dataset ShapeNet~\cite{chang2015shapenet}. Our experiments are conducted on three classes with complex geometry - Car, Table and Chair, containing 7497, 8436 and 6778 shapes respectively. We use the Adam optimizer to optimize the latent code with a step size of $2e-3$. The number of views is $6$ and  hyper-parameters $N(i)$ are set to be $5$. We stop the optimization process when the handle points are $1e-3$ away from its corresponding target points. 

\subsection{Qualitative and Quantitative Evaluation}
\begin{figure}
    \centering
    \includegraphics[width=1 \linewidth]{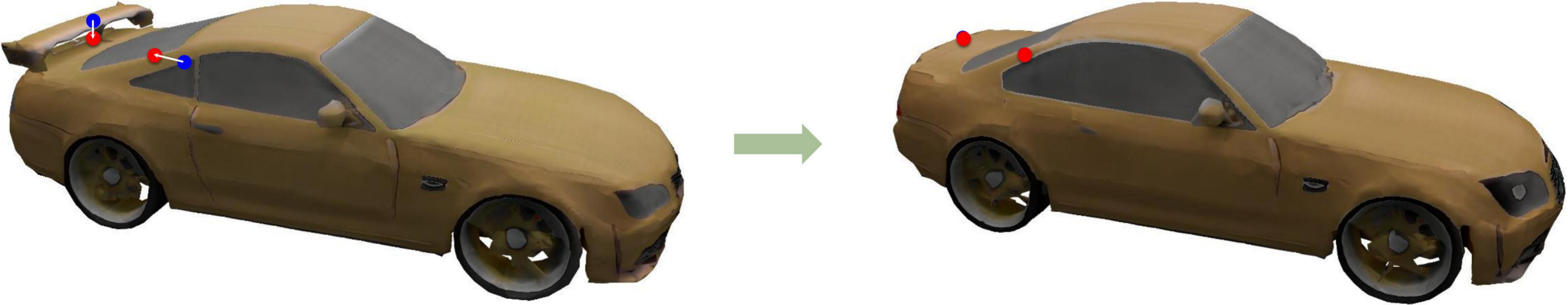}
    \caption{
    Qualitative results of ToW3D. ToW3D achieves precise manipulation when there exists multiple handle points.
    }
    \label{fig:multi}
\end{figure}
In this section, we first qualitatively conduct a qualitative comparative analysis between our proposed approach and Drag3D~\cite{drag3d} which is a vanilla implementation of DragGAN for mesh editing. As illustrated in Fig.~\ref{fig:qualitative_comparison}, we demonstrate the mesh editing results for various object categories and user inputs. Our ToW3D achieves diverse and precise manipulation effects such as changing the shape of various categories and altering the vehicle type while ensuring the preservation of the appearance consistency of original meshes. Although Drag3D is able to move the handle points towards their corresponding target points, it often introduces undesirable appearance changes in the meshes under large deformations. These alterations can include the generation of additional chair arms or the occurrence of broken segments within such object categories. Besides, we also provide the editing result of multiple handle points and their corresponding target points in Fig~\ref{fig:multi}. The qualitative results demonstrate that ToW3D possesses the capability to achieve precise manipulation for multiple handle points and target points simultaneously.

\begin{table}[]

    \centering
     \caption{Quantitative results on the quality of images rendered from edited meshes. We report PSNR, SSIM and LPIPS.}
     \label{tab:quantitative2D}
    \footnotesize
    \begin{tabular}{ccccccc}
        \hline
        Method & \multicolumn{3}{c}{Drag3D} & \multicolumn{3}{c}{ToW3D (Ours)}\\
        Metric &  PSNR$\uparrow$ & SSIM$\uparrow$ & LPIPS$\downarrow$  & PSNR$\uparrow$ & SSIM$\uparrow$ & LPIPS$\downarrow$ \\
        \hline
        Chair & 20.65 & 0.933 & 0.111 & \textbf{24.55} & \textbf{0.977} & \textbf{0.053} \\
        Car   & 23.58 & 0.948 & 0.056 & \textbf{26.84} & \textbf{0.963} & \textbf{0.033} \\
        Table & 13.66 & 0.910 & 0.210 & \textbf{15.76} & \textbf{0.920} & \textbf{0.143} \\
        Average&19.30 & 0.930 & 0.126 & \textbf{22.38} & \textbf{0.953} & \textbf{0.076} \\
        \hline
    \end{tabular}
\end{table}
We then employ five metrics to quantitatively evaluate our method, as illustrated in Table~\ref{tab:quantitative2D} and Table~\ref{tab:quantitative3D}. 
Specifically, we adopt PSNR, SSIM and LPIPS to assess images rendered from a single viewpoint. Higher PSNR and SSIM indicate superior image quality, whereas a lower LPIPS represent higher quality. Additionally, we include FID and Haursdorff Distance (HD) metrics to further measure the quality of the generated mesh. Both metrics should be lower to imply a mesh with higher quality. As demonstrated in Table~\ref{tab:quantitative2D}, our method outperforms Drag3D across various categories in PSNR by 3.08, SSIM by 0.023 and LPIPS by 0.05, thereby denoting that our ToW3D yields a high quality image rendered from a single view. Besides, Table~\ref{tab:quantitative3D} demonstrates that our ToW3D reaches a better result by improving the FID by 83.32 and HD by 1.17, portraying a more realistic mesh compared with Drag3D.

\begin{table}[]

    \centering
    \caption{Quantitative results on the quality of edited meshes compared to original meshes. We report FID and HD.}
    \label{tab:quantitative3D}
    \footnotesize
    \begin{tabular}{ccccc}
        \hline
        Method & \multicolumn{2}{c}{Drag3D} & \multicolumn{2}{c}{ToW3D (Ours)}\\
        Metric &  FID$\downarrow$ & HD$\downarrow$ & FID$\downarrow$  & HD$\downarrow$ \\
        \hline
        Chair & 142.60 & 1.19 & \textbf{47.67} & \textbf{1.00}  \\
        Car   & 169.79 & 1.21 & \textbf{56.76} & \textbf{0.48}  \\
        Table & 114.29 & 5.49 & \textbf{72.30} & \textbf{2.89}  \\
        Average&142.23 & 2.63 & \textbf{58.91} & \textbf{1.46}  \\
        \hline
    \end{tabular}
\end{table}

\subsection{Ablation Study}

In this part, we conduct the ablation study on our proposed modules in our framework to evaluate their effectiveness.  We develop the following variants: (a) the baseline method comprised of motion supervision and point tracking, (b) the baseline method with structure adaption module, (c) the baseline method with semantic preservation module and our ToW3D method. 
The quantitative results are listed in Table~\ref{tab:ablation}. 
We observe that variant (b) solely employing structure adaption module and variant (c) exclusively adopting semantic preservation module significantly enhance the quality of edited mesh , which demonstrate the effectiveness of both modules. The best performance is achieved when combining structure adaption module and semantic preservation module. 
\begin{table}[]
    \centering
    \caption{Effectiveness of our proposed structure adaption module (SA), semantic preservation module (SP) and DynDistill module (DD).}
    \label{tab:ablation}
    \footnotesize
    \begin{tabular}{cccccccc}
        \hline
        ID &SA &SP &  PSNR$\uparrow$ & SSIM$\uparrow$ & LPIPS$\downarrow$ & FID$\downarrow$ & HD$\downarrow$\\
        \hline
        (a)   & -           & -           & 12.59 & 0.905 & 0.179 & 141.58 & 7.7 \\
        (b)   &$\checkmark$ & -           & 14.02 & 0.922 & 0.094 & 54.25 & 2.4 \\
        (c)   &-            &$\checkmark$ & 15.61 & 0.933 & 0.082 & 44.11  & 3.2 \\
        ToW3D &$\checkmark$ &$\checkmark$ & \textbf{17.09} & \textbf{0.949} & \textbf{0.070} & \textbf{28.69} & \textbf{1.5} \\
        \hline
    \end{tabular}
\end{table}

\section{Conclusion}
In this work, we propose a novel mesh editing method named ToW3D, which focuses on precise and consistent control over 3D Generative Adversarial Networks (GANs) with the Tug-of-War competition between shape deformation and appearance consistency. To avoid drastic changes in global appearance when editing local areas of meshes, ToW3D performs two optimization steps iteratively, following the strategy of “drag locally, shove globally”. The first optimization step incrementally guides handle points towards their target locations while the second optimization step aims to direct the structure and semantics to align with the original mesh for consistency. Extensive qualitative and quantitative experiments show the effectiveness and superiority of our method regarding appearance consistency and fidelity especially under large deformations.

\section*{Acknowledgment}

This work was supported in part by the National Natural Science Foundation of China under Grant 62206147.

\bibliographystyle{IEEEbib}
\bibliography{main}

\end{document}